\documentclass[11pt,a4paper]{article}
\usepackage[hyperref]{acl2020}
\usepackage{times}
\usepackage{latexsym}
\usepackage{graphicx}
\usepackage{multirow,amsmath,mathrsfs}
\usepackage{float}
\usepackage{enumitem}
\usepackage{microtype}
\graphicspath{{images/}}

\aclfinalcopy 

\makeatletter
\newcommand\footnoteref[1]{\protected@xdef\@thefnmark{\ref{#1}}\@footnotemark}
\makeatother
\title{A Deep Learning System for \\Sentiment Analysis of Service Calls}

\author{Yanan Jia \\
    Businessolver /  Bellevue, WA \\
    \texttt{jia.66@osu.edu} \\\And
    Sony SungChu \\
    Businessolver / Bellevue, WA \\
    \texttt{susungchu@businessolver.com} \\}

\date{}

\begin{document}
    \maketitle
    \begin{abstract}
Sentiment analysis is crucial for the advancement of artificial intelligence (AI).
Sentiment understanding can help AI to replicate human language and discourse.  Studying the formation and response of sentiment state from well-trained Customer Service Representatives (CSRs) can help make the interaction between humans and AI more intelligent. 

In this paper,  a sentiment analysis pipeline is first carried out with respect to real-world multi-party conversations--that is, service calls. Based on the acoustic and linguistic features extracted from the source information,  a novel aggregated method for voice sentiment recognition framework is built. Each party's sentiment pattern during the communication is investigated along with the interaction sentiment pattern between all parties.        
        
    \end{abstract}
    
\section{Introduction}
The natural reference for AI systems is human behavior. In human social life, emotional intelligence is important for successful and effective communication.  A human has the natural ability to comprehend and react to the emotion of their communication partners through vocal and facial expressions  \cite{Margarita:2012, soujanya:2015}. 
A long-standing goal of AI has been to create affective agents that can recognize, interpret and express emotions. 

Early-stage research in affective computing and sentiment analysis has mainly focused on understanding affect towards entities such as movie, product, service, candidacy, organization, action and so on in monologues, which involves only one person's opinion.  
However, with the advent of Human-Robot Interaction (HRI) such as voice assistants and customer service chatbots, researchers have started to build empathetic dialogue systems to improve the overall HRI experience by adapting to customers' sentiment.

Sentiment study of Human-Human Interactions (HHI) can help machines identify and react to human non-verbal communication which makes the HRI experience more natural. 
The call center is a rich resource of communication data. A large number of calls are recorded daily in order to assess the quality of interactions between CSRs and customers. 
Learning the sentiment expressions from well-trained CSRs during communication can help AI understand not only what the user says, but also how he/she says it so that the interaction feels more human.

In this paper,  we target and use real-world data -- service calls, which poses additional challenges with respect to the artificial datasets that have been typically used in the past in multimodal sentiment researches, such as variability and noises.  The basic `sentiment' can be described on a scale of approval or disapproval, good or bad, positive or negative, and termed polarity \cite{soujanya:2014}.  

In the service industry, the key task is to enhance the quality of services by identifying issues that may be caused by systems, rules, or service qualities. These issues are usually expressed by a caller's anger or disappointment on a call. In addition,  service chatbots are widely used to answer customer calls. If customers get angry during HRI,  the system should be able to transfer the customers to a live agent. 
In this study, we mainly focused on identifying `negative' sentiment, especially `angry' customers.   
Given the non-homogeneous nature of full call recordings, which typically include a mixture of negative, and nonnegative statements, sentiment analysis is addressed at the sentence level. Call segments are explored in both acoustic and linguistic modalities. The temporal sentiment patterns between customers and CSRs appearing in calls are described.

The paper is organized as follows: Section \ref{se:relatedwork} covers a brief literature review on sentiment recognition from different modalities; Section \ref{se:piplineanddata} proposes a pipeline which features our novelties in training
data creation using real-world multi-party conversations, including a description of the data acquisition, speaker diarization, transcription, and semi-supervised learning annotation; the methodology for acoustic and linguistic sentiment analysis are presented in Section \ref{se:models}; Section \ref{se:fusion} illustrates the methodology adopted for fusing different modalities;  Section \ref{se:results} presents experimental results including the evaluation measures and temporal sentiment patterns; finally,  Section \ref{se:discussion} concludes the paper and outlines future work.

\section{Related Work}\label{se:relatedwork}
In this section, we provide a brief overview of related work about text-based and audio-based sentiment analysis.

\subsection{Text-based Sentiment Analysis}
Sentiment analysis has focused primarily on the processing of text and mainly consists of either rule-based classifiers that make use of large sentiment lexicons, or data-driven methods that assume the availability of a large annotated corpora.

Sentiment lexicon is a list of lexical features (e.g. words) which are generally labeled according to their semantic orientation as either positive or negative \cite{Liu10}.  Widely used lexicons include binary polarity-based lexicons such as Harvard General Inquirer \cite{stone:1966}, Linguistic Inquiry and Word Count (LIWC, pronounced `Luke')  \cite{pennebaker2001, pennebaker2007},  
Bing \cite{Liu2012SentimentAA}, and valence-based lexicons, such as AFINN \cite{AFINN2011}, SentiWordNet \cite{sentiwordnet2013}, and SnticNet \cite{senticnet2010}. 
Employing these lexical, researchers can apply their own rules or use existing rule-based modeling, such as VADER \cite{hutto:2015} to do sentiment analysis. 

One big advantage for the rule-based model is that these approaches require no training data and generalizes to multiple domains. However, since words are annotated based on their context-free semantic orientation, word-sense disambiguation \cite{hutto:2015} may occur when the word has multiple meanings. For example, words like `defeated', `envious',  and `stunned'  are classified as `positive' in Bing, but `-2' (negative) in AFINN. 
Although the rule-based algorithm is known to be noisy and limited, a sentiment lexicon is a useful component for any sophisticated sentiment detection algorithm and is one of the main resources to start from \cite{soujanya:2014}.

Another major line of work in sentiment analysis consists of data-driven methods based on a large dataset annotated for polarity. The most widely used datasets include the MPQA corpus which is a collection of manually annotated news articles \cite{Janyce:2005, Wilson:2005}, movie reviews with two polarity \cite{Bo:2004}, a collection of newspaper headlines annotated for polarity \cite{Carlo:2007}.  With a large annotated datasets, supervised classifiers have been applied  \cite{go:2009, Socher:2013}.
Such approaches step away from blind use of keywords and word co-occurrence count, but rather rely on the implicit features associated with large semantic knowledge bases \cite{cambria:2015}.

\subsection{Audio-based Sentiment Analysis}
Vocal expression is a primary carrier of affective signals in human communication. 
Speech as signals contains several features that can extract linguistic, speaker-specific information, and emotional. Related work about audio-based sentiment analysis along with multimodal fusion is reviewed in this section. 

Studies on speech-based sentiment analysis have focused on identifying relevant acoustic features. 
Use open source software such as OpenEAR \cite{eyben:2009},  openSMILE \cite{eyben:2010},  JAudio toolkit \cite{mcennis:2005} or library packages \cite{mcfee:2015, sueur:2008} to extract features.  
Those features along with some of their statistical derivates are closely related to the vocal prosodic characteristics, such as a tone, a volume, a pitch, an intonation, an inflection, a duration, etc.  

Supervised or unsupervised classifiers can be fitted based on the statistical derivates of those features \cite{jain:2018, pan:2012}. Sequence models can be fitted based on filter banks, Mel-frequency cepstral coefficients (MFCCs), or other low-level descriptors extracted from raw speech without feature engineering \cite{aguilar2019}. However, this approach usually requires highly efficient computation and large annotated audio files.

Multimodal sentiment analysis has started to draw attention recently because of the unlimited multimodality source of information online, such as videos and audios \cite{soujanya:2016, poria:2015, erik:2017}.  Most of the multimodal sentiment analysis today is  focused on monologue videos.  In the last few years, sentiment recognition in conversation has started to gain research interest, since reproducing human interaction requires a deep understanding of the conversation, and sentiment plays a pivotal role in conversations.  The existing conversation datasets are usually recorded in a controlled environment, such as a lab, and segmented into utterances, transcribe to text and annotated with emotion or sentiment labels manually. Widely used dataset includes AMI Meeting Corpus \cite{Carletta:2005}, IEMOCAP \cite{Busso:2008a},  SEMAINE \cite{Mckeown:2013} and AVEC \cite{Schuller:2012}. 

Recently, a few recurrent neural network (RNN) models are developed for emotion detection in conversations, e.g.  DialogueRNN \cite{Majumder:2019} or ICON\cite{Hazarika:2018a}. 
However  they are less accurate in emotion detection for the utterances with emotional shift \cite{poria:2019} and the training data requires the speaker information. The conversation models are not employed in our polarity sentiment analysis because of the quality of the data and the approach used to gain the training data. More detailed explanations can be found in Section \ref{subse:activelearning}.

At the heart of any multimodal sentiment analysis engine is the multimodal fusion. The multimodal fusion integrates all single modalities into a combined single representation. Features are extracted from the data from each modality independently. Decision-level fusion feeds the features of each modality into separate classifiers and then combines their decisions.  Feature-level fusion concatenates the feature vectors obtained from all modalities and feeds the resulting long vector into a supervised classifier.
Recent research on multi-modal fusion for sentiment recognition has been conducted at either the feature level or decision level. 
    

    \section{Dataset and Pipeline}\label{se:piplineanddata}
    The data resources used for our experiments are described in Section \ref{subse:data}. Data preparation including speech transcription and speaker diarization is discussed in Section \ref{subse: dataprep}. The sentiment annotation guideline is introduced in Section \ref{subse: annotation}. Section \ref{subse:activelearning} presents a semi-supervised learning annotation pipeline that chains data preparation, model training, model deploying and data monitor. 
    
    \subsection{BSCD: Benefits Service Call Dataset}\label{subse:data}
    The main dataset we created in this paper consists of service calls collected from a health care benefits Call Center (named BSCD). Calls are focused on customers looking for help or support with company provided benefits such as health insurance. 
    
    500 calls are collected from the call center database covering diverse topics, such as insurance plan information, insurance id card, dependent coverage, etc. The call data set had female and male speakers randomly selected with their age ranging approximately from 16-80. Calls involving translators are eliminated to keep only speakers expressing themselves in English. All the calls are presented in Wave format with a sample rate of 8000 Hertz and duration varying from 4 minutes to 26 minutes. All calls are pre-processed to eliminate repetitive introductions. The beginning of each call contains an introduction of the users' company name by a robot. To address this issue, the segment before the first pause (silence duration $>$ 1 second) is removed from each call. 
    
    A robust computational model of sentiment analysis needs to be able to handle real-world variability and noises. While the previous research on multimodal sentiment or emotion analysis used audio and visual recorded in laboratory settings \cite{Busso:2008a, Mckeown:2010, Mckeown:2013}; the BSCD gathered real-world calls contains ambient noise present in most audio recordings, as well as diversity in person-to-person communication patterns. Both of these conditions result in difficulties that need to be addressed in order to effectively extract useful data from these sources.

\subsection{Data Preparation}\label{subse: dataprep}
To discard noise and long pauses (silence duration $>$ 5 seconds) in calls, Voice Activity Detection (VAD) is applied,  followed by the application of Automatic Speech Recognition (ASR) and Automatic Speaker Diarization (ASD) to transcribe the verbal statements, extract the start and end time of each utterance, and identify the speaker of each utterance. Each call is segmented into an average of 69 utterances. The duration of the utterances is right-skewed with a median of 2.9 seconds;  first and third quantiles 1.6 and 5.1 seconds.  

\begin{figure}
    \centering
    \includegraphics[scale=0.28]{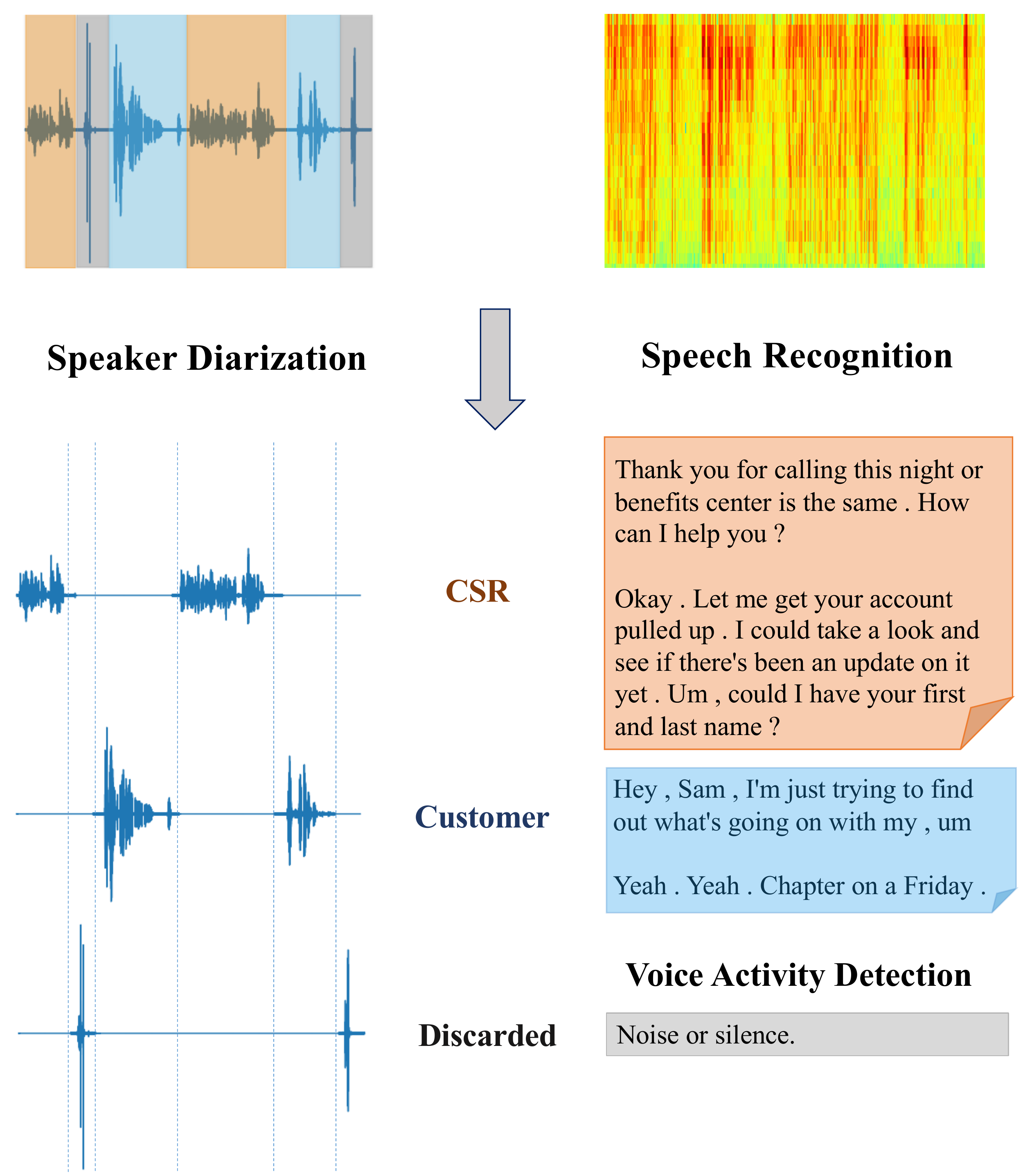}
    \caption{Data preparation workflow}
    \label{fig:dataprep}
\end{figure}

By searching keywords such as `How can I help' in the content of each utterance, speakers are labeled as CSR or customer. Each utterance is linked to the corresponding audio stream, auto transcription, as well as speaker label. 
The workflow and corresponding results for the first 23 seconds of one selected call are shown in Figure \ref{fig:dataprep}, where the original input is a call audio sample. After data preparation, segments of noise and silence are discarded. This call sample is segmented into 4 utterances. The audio streams are from the original audio and split based on the start and end time of each utterance. Auto transcriptions are more likely to be ungrammatical if the recording quality is bad or the conversation contains words that ASR cannot identify or the speakers do not express themselves clearly. 
The ungrammatical transcriptions usually occur in customer parts and the frequency of ungrammaticality varies from case to case. Although the sentiment recognition of a whole call tends to be robust with respect to speech recognition errors, the sensitivity of each utterance analysis to ASR errors is not reparable given our study. 
The Speaker labels are from ASD output which can be misclassified because of the occurrence of speakers overlapping or speakers with similar acoustic features. 
Conversation sentiment pattern study can be misleading due to the misclassified ASD output, although misclassified ASD is rare.

This process allows us to study features from both modalities: transcribed words and acoustics. 
Distinguishing different parties gives us the ability to study the temporal sentiment transitions of individual speakers and interactions among speakers in a conversation.
However, since the data preparation is part of the pipeline described in section \ref{subse:activelearning}, which runs in real-time, sentiment analysis must rely on error-prone ASR and ASD outputs.

\subsection{Sentiment Annotation}\label{subse: annotation}
Sentiment annotation is a challenging task as the label depends on the annotators' perspective,  and the differences inherent in the way people express emotions. The sentiment is opinion-based, not fact-based. 
This study aims at identifying negative expressions in calls, especially angry customers who are not satisfied with the services and the business or system rules. By identifying and studying those types of cases, the business can improve call center services and fix the possible business or system issues. 

Guidelines are set up for the annotation. The customer negative tag is for negative emotions  (e.g. ``I hate the system"), attitudes (e.g.  ``I am not following you"), evaluations (e.g.  ``your service is the worst"), and negative facts caused by other parties (e.g.  ``I never received my card"). Other negative facts are not considered as negative (e.g. ``My wife died, I need to remove her from my dependents").  The guidelines for CSRs are different. Well trained CSRs usually do not respond negatively, but there are cases that they cannot help the customers. We identify those cases as negative. Cases where a CSR cannot help the customer usually involve business process or system issues.

The sentiment is not always explicit in the text. Borderline linguistic utterances stated loudly and quickly are usually identified as `negative'. E.g. the utterance ``Trust me, it could be done" is classified as negative, since it is in the context that the representative fails to help the customer to enroll in the health plan, and in the audio, the customer is irritated. In all the multimodal sentiment analysis,  the labels of all modalities are kept consistent for the same utterance. In our data annotation process, we also keep both text and audio labels that agree with each other and the annotation is based on the audio segments.

 \subsection{Semi-supervised Learning Annotation Pipeline}\label{subse:activelearning}
 
 \begin{figure*}
     \centering
     \includegraphics[scale=0.35]{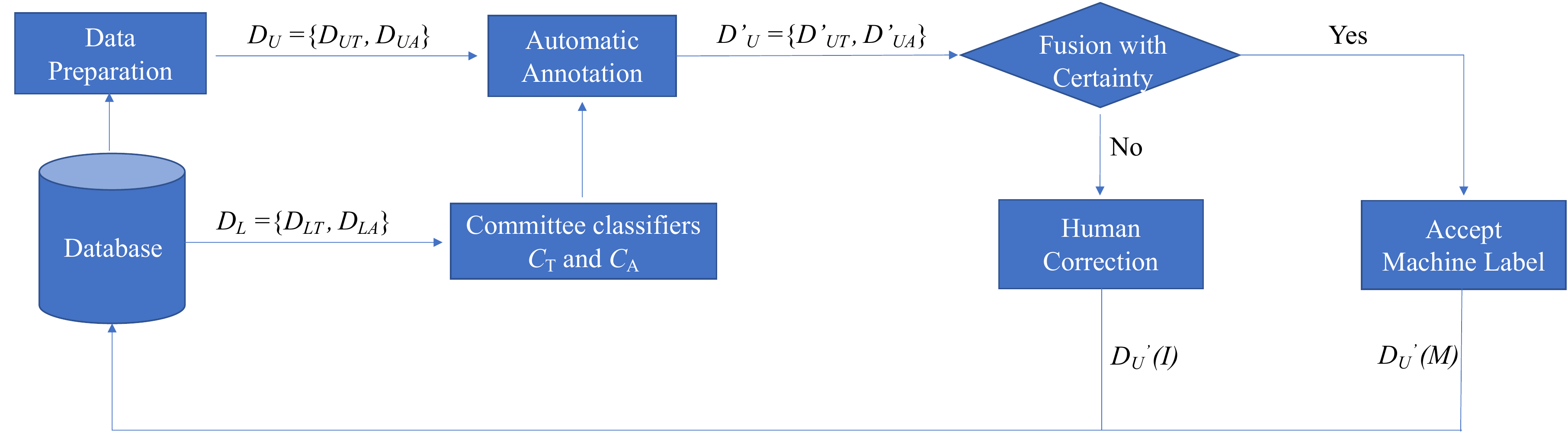}
     \caption{Semi-supervised learning annotation pipeline}
     \label{fig:activelearning}
 \end{figure*}
 
 To successfully run and train analytical models, massive quantities of stored data are needed. Creating large annotated datasets can be a very time consuming and labor-intensive process. To keep the human annotation effort to a minimum, a semi-supervised learning annotation scheme is applied to tag the polarity of utterances as negative, or nonnegative. Figure  \ref{fig:activelearning}  illustrates the process which is similar to active learning annotation. It takes as input a set of labeled examples $D_{L}$  including text $D_{LT}$ and audio $D_{LA}$, as well as a larger set of unlabeled examples $D_{U}  = \{ D_{UT} , D_{UA}  \}$, and produces committee classifiers $C  = \{ C_{T} , C_{A}  \}$ and a relatively small set of newly labeled data  $D_U^{\prime}(I)$ and $D_U^{\prime}(M)$ \cite{olsson:2009}. 
 
 Semi-supervised learning annotation cooperates with humans and machines and combines both semi-supervised learning and multiple classifiers approach for corpus annotation. This pipeline consists of several steps: data generation to obtain $D_{U}$ (Section \ref{subse: dataprep}), model training for both modalities to obtain $C_{T}$ and $C_{A}$  (Section \ref{se:models}), model deployment to get machine label $D_{U}^\prime = \{ D_{UT}^\prime, D_{UA}^\prime \}$, model fusion (Section \ref{se:fusion}) and results evaluation to decide whether to accept machine label $D_U^{\prime}(M)$ or ask a human annotator  for classifications of the utterances to obtain  $D_U^{\prime}(I)$, then move $D_U^{\prime}(I)$ and $D_U^{\prime}(M)$ from $D_U^\prime$ to $D_L$. It is cyclical and iterative as every step is repeated to continuously improve the accuracy of the classifier and achieve a successful algorithm. 
 
 Note, the classifiers in committee  $C  = \{ C_{T} , C_{A}  \}$ are modified based on $D_L$ in each iteration. 
 The annotation process starts with 20 calls selected from the service center by human domain experts, 20 calls are chunked to 1410 segments via data preparation processing and annotated by three annotators manually as $D_L$.
 For the first three iterations, set $C_T$=\{Support Vector Machine (SVM), VADER, AWSSA\footnote[1]{AWS Comprehend Sentiment Analysis API},   AWSCC\footnote[2]{AWS Custom Classification API}, GoogleSA\footnote[3]{Google Language Sentiment Analysis API}\} requires a small size of training data or no extra training data. As the size of $D_{LT}$ increases, we form a new committee $C_T$ = \{SVM,   Long Short-Term Memory (LSTM),  Bidirectional Long Short-Term Memory (BLSTM)\}. These classifiers are described in Section \ref{se:text}.  Section \ref{subse:audio} introduced $C_A$ = \{Elastic-Net Regularized Generalized Linear Models (Elastic-Net), K-Nearest Neighbors (KNN), Random Forest (RF),  Gaussian Mixture Model (unsupervised GMM) \}. In the later iterations, Recurrent Neural Networks (RNN) such as LSTM and BLSTM are applied. 
 
 If one call has a long duration ($T>10$ minutes) and a high percentage of negative utterances based on $D_{U}^{\prime} $ ($>40\%$ for customer or $>20\%$ for CSR),  then we say this call is potentially negative and informative.  We then ask an annotator to manually correct the annotated tags $D_{U}^{\prime}$  by listening to the call, and move the results $D_{U}^{\prime}(I)$ to  $D_{L}$. For all the other calls, we only keep the utterances where classifiers all agree as $D_{U}^{\prime}(M)$.  We then remove chunks that are  too short (duration $<$ 1s) or too long (duration $>$20s). Finally, we discard chunks where the annotator cannot discern classification.
 
 Using the pipeline,  6,565 negative and 10,322 nonnegative call clips were annotated as the training dataset. 
 The training data $D_{LT}$ still include transcription errors, even though the threshold discussed in the above paragraph is set to eliminate those utterances to add to the training dataset. 
 In addition, 18,705 cleaned text chat data collected from chat windows are also added to $D_{LT}$ via the annotation pipeline to improve the  $C_T$ accuracy, the details are shown in Section \ref{se:resultsmeasure}. 
 
 Because of the quality of the calls, the poor performance of the ASR for some cases, and the threshold used to annotate the utterances, more than half of the original call segments are discarded\footnote{The accuracy on the test data decreases 8\% by including all the call segments in the training dataset} and 18,705 text chat data are added to  $D_{LT}$=\{transcription data,  \emph{chat data}\} without the corresponding audio files in  $D_{LA}$.  It is hard to consider the context of the conversation since the segments are not continuing in the training dataset. Therefore,  conversation models are not considered in our committee classifiers $C$.    

    \section{Bimodal Sentiment Analysis}\label{se:models}
To model information for sentiment analysis from calls,  we first obtain the streams corresponding to each modality via the methods described in Section \ref{subse: dataprep}, followed by the extraction of a representative set of features for each modality.  These features are then used as cues to build a classifier of binary sentiment.

\subsection{Sentiment Analysis of Textual Data}\label{se:text}
General approaches such as sentiment lexical and sentiment APIs are easy to apply. Both approaches are employed in $C_T$ to monitor the utterance prediction labels in the early stage of semi-supervised learning annotation to extend training data. 

VADER \cite{hutto:2015} is a simple rule-based model for general sentiment analysis. The results have four categories: compound, negative, neutral, and positive. 
We classify utterances with negative output as negative,   neutral and positive as nonnegative\footnote{\label{note1}. Utterances with compound or mixed class are very few, and they are discarded to keep the training data clear} so that it is consistent with BSCD annotation. 
This model has many advantages, such as being less computationally expensive and easily interpretable. 
However, one of the main issues with only using lexicons is that most utterances do not contain polarized words. The utterances without polarized words are usually classified as neutral or nonnegative\footnote{The high Rec(+) and low Rec(-) shown in table \ref{table:textresults} verifies this conclusion.}.

Sentiment analysis API is another way to classify sentiment without extra training data. Amazon offers Sentiment Analysis in Amazon Comprehend (AWSSA), which uses machine learning to find insights and relationships in a text.  The result returns Mixed, Negative, Neutral, or Positive classification. 
To be consistent with the BSCD we created, Neutral and Positive are combined as one classifier: nonnegative\footnoteref{note1}.
Another sentiment analysis on Google Cloud Natural Language API (GoogleSA) also performs sentiment analysis on text. Sentiment analysis attempts to determine the overall attitude and is represented by numerical score and magnitude values. We simply set utterances with negative scores as negative and nonnegative otherwise.

For machine learning-oriented techniques by linguistic features, we evaluated well-known SVM, LSTM, and BLSTM models. Since the data is unbalanced and we want the model to focus more on the negative class,  we apply weighted loss functions during the training. 
Hyperparameters are tuned for each model, and ensemble models are also developed by taking the weighted majority vote.  

\begin{table*}
    \centering
    \resizebox{0.9\textwidth}{!}{
        \begin{tabular}{l|ccc|ccc|ccccc} 
            \hline
            \multirow{2}{*}{Methods} & \multicolumn{6}{c}{Text}      & \multicolumn{5}{|c}{Audio}      \\ \cline{2-12} 
            & SVM         & LSTM        & BLSTM            & Vader  & AWS     & Google  & Elastic-Net   & KNN  & RF & GMM & BLSTM  \\ \hline
            Acc.      & 0.814 &0.853&0.843&  0.498 & 0.651  &  0.637 & 0.570  & 0.544  &0.585 & 0.546 & 0.601   \\ \hline
            F1 (w)         & 0.814   & 0.852     & 0.842    &  0.347    & 0.628         & 0.615    &0.500   & 0.534  &0.549 & 0.500 &0.584    \\ \hline
            Prec.(+)    &  0.770& 0.802 &0.781  &  0.494 &0.594 & 0.586  & 0.528   &  0.518  & 0.541 & 0.513 &0.561 \\ \hline
            Prec.(-)   &0.871& 0.92& 0.934 &  0.742 & 0.821& 0.779      &0.860      &   0.589  & 0.741 & 0.685  &0.693 \\ \hline
            Rec. (+)    & 0.886 &0.929& 0.946 &  0.991& 0.908&  0.881  &  0.964 &  0.697&  0.883 &  0.872  &0.811\\ \hline
            Rec. (-)  &0.746 & 0.779 &  0.745   &   0.024 & 0.404&0.402    & 0.205  & 0.402 &0.309   & 0.252   & 0.402\\ \hline
    \end{tabular}  }
    \caption{Binary classification of sentiment polarity on test data: Accuracy (Acc.), weighted F1-score (F1 (w)),  precision (Prec.) and recall (Rec.) for the nonnegative (+) and negative (-) classes.}
    \label{table:textresults}
\end{table*}

 \subsection{Sentiment Analysis of Acoustic Data}\label{subse:audio}

Feature engineering heavily relies on expert knowledge about data features. 
To better understand the human hearing process, we study the acoustic features based on human perception. 
Three perceptual categories are described in this section.  Their corresponding features are usually short-term based features that are extracted from every short-term window (or frame). Long-term features can be generated by aggregating the short-term features extracted from several consecutive frames within a time window. For each short-term acoustic feature, we calculated nine statistical aggregations: mean, standard deviation,  quantiles (5\%, 25\%, 50\%, 75\%, 95\%), range (95\%-5\% quantile), and interquartile range (75\%-25\% quantile) to get the long-term features of each segment. 

\vspace{-0.3em}
\begin{itemize}[wide=0pt]
    \setlength\itemsep{-0.5em}
    \item \textbf{Loudness} is the subjective perception of sound pressure which is related to sound intensity. Amplitude and mean frequency spectrum features are extracted to measure loudness. The greater the amplitude of the vibrations, the greater the amount of energy carried by the wave, and the more intense the sound will be. 
    \item \textbf{Sharpness} is a measure of the high-frequency content of a sound, the greater the proportion of high frequencies the ‘sharper’ the sound.  Fundamental frequency (pitch) and dominant frequency are extracted.  
    \item \textbf{Speaking rate} is normally defined as the number of words spoken per minute. In general, the speaking rate is characterized by different parameters of speech such as pause and vowel durations. 
    In our study, speaking rate is measured by pause duration, character per second (CPS), and word per second (WPS) which are calculated as following for the  $i$th segment:   
    \[
    \text{Pause duration}_i= \frac{T^{silence}_i}{T^{total}_i}  
    \]
    \[
    \text{CPS}_i= \frac{N^{character}_i}{T^{total}_i},   \hspace{1cm}    \text{WPS}_i= \frac{N^{word}_i}{T^{total}_i} 
    \]
    where  for segment $i$,  $T_i$ denotes the time,  and $N_i$ denotes the number of characters or words in the corresponding transcription.
    Pause duration can be interpreted as the percentage of the time where the speaker was silent. The three variables are aggregated statistics, long-term features.  
    
\end{itemize}
\vspace{-0.2em}
In nonnegative cases,  speakers are in a relaxed and normal emotional state. An agitated or angry emotional state speaker will typically be characterized by increased vocal loudness, sharpness, and speaking rate. $C_A$ =\{Elastic-Net, KNN, RF, GMM\} are built based on the 39 selected features. 
    

    “Hand-crafted” features are generally very successful for specificity sound analysis tasks. One of the main drawbacks of feature engineering is that it relies on transformations that are defined beforehand and ignore some particularities of the signals observed at runtime such as recording conditions and recording devices. A more common approach is to select and adapt features initially introduced for other tasks. A now well-established example of this trend is the popularity of MFCC features \cite{serizel2017}. In our experiments, MFCC is extracted from each segment and fed to RNN models in later iterations with $|D_{LA}|>10,000$. 
  
  \begin{table}
      \centering
      \resizebox{0.45\textwidth}{!}{
          \begin{tabular}{l|ccc|cc} 
              \hline
              \multirow{2}{*}{Methods} & \multicolumn{3}{c|}{Ensemble}   & \multicolumn{2}{c}{Fusion}         \\ \cline{2-6} 
              &Text    & Audio     & T+A       &  Fus1  &Fus2            \\ \hline
              Acc.     & 0.851   & 0.586& 0.846   & 0.858 & 0.871\\ \hline
              F1 (w)      &  0.851     & 0.525  &0.846 &   0.858  & 0.871        \\ \hline
              Prec.(+)    &  0.779 &0.531& 0.800  &0.790 &0.818\\ \hline
              Prec.(-)    & 0.949 &0.927  &0.896  &0.946  & 0.933      \\ \hline
              Rec. (+)    &0.953 & 0.979&0.894 &0.950   &0.933 \\ \hline
              Rec. (-)   &  0.761& 0.240 & 0.804 &0.777 &0.817 \\ \hline
      \end{tabular}}
      \caption{Binary classification of sentiment polarity on both linguistic and acoustic modalities.}
      \label{table:fusresults}
  \end{table}

    \section{Fusion }\label{se:fusion}
 There are two main fusion techniques: feature-level fusion and decision-level fusion. In our experiments, we employ decision-level fusion. 
 Decision-level fusion has many advantages \cite{poria:2015}. One benefit of the decision-level fusion is we can use classifiers for text and audio features separately. The unimodal classifier can use data from another communication channel  of the same type to improve its accuracy, e.g. text data from the chat window is borrowed to improve the $C_T$ accuracy in our study. 
 Separating modalities permit us to use any learner suitable for the particular problem at hand. 
 Another benefit of the decision-level fusion is its processing speed since fewer features are used for each classifier and separate classifiers can be run in parallel.
 
 Decision-level fusion usually adds probabilities or summarized prediction from each unimodal classifier with weights or takes the majority voting among the predicted class labels by unimodal classifiers.

In this paper, various fusion methods are evaluated, including a novel approach that uses linguistic ensemble results as the baseline, while then checking acoustic results to modify classification decisions. In Fus1, if the audio ensemble classifies negative and one or more text models classifies negative, we then reclassify the result to negative. In Fus2, if the audio ensemble classifies a sample as negative, we then reclassify the result to negative directly without checking the linguistic modality. 
 
 \section{Experiment Results}\label{se:results}
 The test dataset consists of 21 calls with 1,890 utterances, which are manually  annotated  for negative (848)  and nonnegative (1,042). 
 
 \begin{figure*}
     \centering
     \includegraphics[scale=0.4]{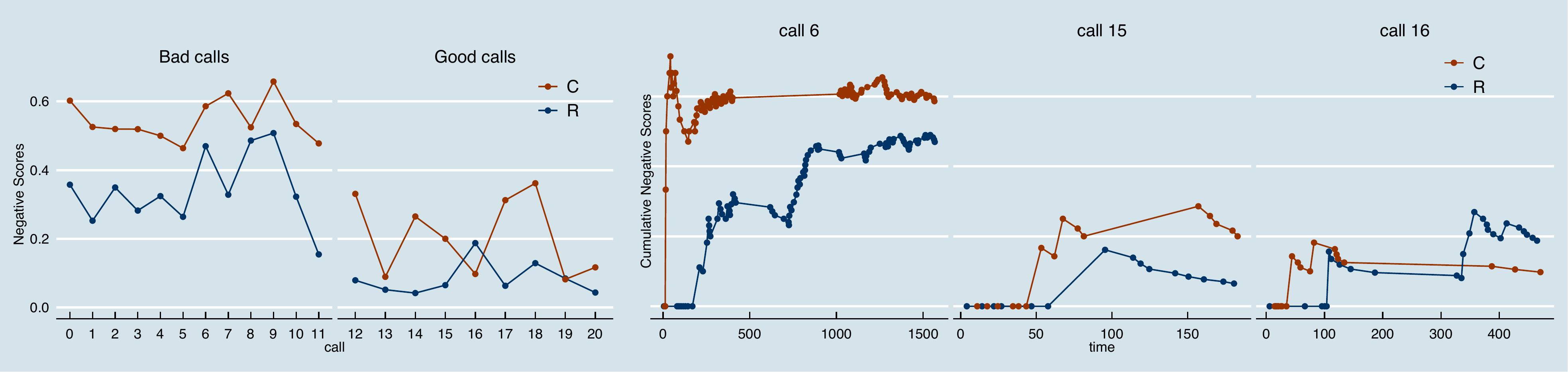}
     \caption{The (cumulative) negative score pattern between  customers  and CSRs}
     \label{fig:cumnegscore}
 \end{figure*}
 \subsection{Evaluation Measures}\label{se:resultsmeasure}
 As evaluation measures, we rely on accuracy and weighted F1-score,
 which is the weighted harmonic mean of precision and recall.  Precision is the probability of returning values that were correct. Recall, also known as sensitivity is the probability of relevant values that the algorithm outputs.

 As shown in Table \ref{table:textresults}, 
 general approaches in $C_T$, Vader and APIs,  tend to have a low negative recall.  The semantic knowledge based classifiers have more than 20\%  higher  F1-score than the general approaches. The classifiers are trained on $D_{LT}$=\{transcription data, chat data\}. The overall F1-score is more than 10\% higher than the classifiers trained on call transcription only data. 
 
 BLSTM  on MFCC performs better than $C_A$ = \{Elastic-Net (penalty  $0.2 ||\beta||_1 + 0.4 ||\beta||_2^2$), KNN ($k=3$), RF, GMM\} on  acoustic features.
 Using audio features alone, an F1-score of 0.584 can be reached, which is acceptable considering that the real world audio-only system exclusively analyzes the tone of the speaker's voice and doesn't consider any language information.

 The acoustic modality is significantly weaker than the linguistic modality.
 In most cases, text already includes enough information to judge the sentiment. A few observed typical situations leading to linguistic modality misclassification are the presence of misleading linguistic cues, 
 ambiguous linguistic utterances whose sentiment polarity are highly dependent on the context described in earlier or later part of the call, or nonnegative linguistic utterances stated angrily. 
 In order to achieve better accuracy, we combine the two modalities together to exploit complementary information.
 
 We simply combine results of the three semantic knowledge based classifiers and all the five audio classifiers by taking the weighted majority vote. The T+A ensemble results are shown in Table \ref{table:fusresults} and they do not improve when compared to the unimodal text ensemble results.

 Since the unimodal performance of linguistic modality is notably better than acoustic modality, our decision-level fusion methods use linguistic ensemble results as the base-line, while acoustic results are used as supplemental information to calibrate each classification.
 The Fus2 bimodal system discussed in Section \ref{se:fusion} yields a 2\% improvement in F1-score than the text unimodal system.
 
 The acoustic modality provides important cues to identify negative emotions. It can help correct misclassified nonnegative/ambiguous linguistic utterances.  
 Our results show that relying on the joint use of linguistic and acoustic modalities allows us to better sense the sentiment being expressed as compared to the use of only one modality at a time. 
 The acoustic feature analysis helps us to better understand the spoken intention of the speaker, which is not
 normally expressed through text.

    \subsection{Tempo Sentiment Pattern} \label{se:pattern}

The sentiment is not only regarded as an internal psychological phenomena but also interpreted and processed communicatively through social interactions. 
Conversations exemplify such a scenario where inter-personal sentiment influences persist. 

The left panel in Figure \ref{fig:cumnegscore} shows the negative scores of CRSs and customers in the 21 test calls.  The negative score,  a weighted negative segment percentage,  is calculated to analyze the overall sentiment.  Weights 0.8, 1, and 1.2 are assigned to the first third, second third and last third of each call. 
The negative scores of CRSs are usually lower than customers', and usually high negative scores for customers correspond to high negative scores for CSRs.  
We can conclude from the figure that sentiment can be affected by other parties during a conversation. 

To further analyze the interactions between customers and CSRs, the cumulative negative scores for call 6, 15, and 16 are drawn on the right panel of Figure \ref{fig:cumnegscore}. Call 6 shows the sentiment patterns of a typical bad call, which is characterized by long duration and long hold. The customer has a high negative score from beginning to end, and the CSR fails to help the customer during the call. Call 15 is a typical good call. The overall negative score is low and the negative score pattern goes down for both the customer and the  CSR, which means the problem is resolved by the end of the call. Call 16 is another type of call, in which the customer does not get angry even though the CSR is unable to solve his/her issues.

\section{Discussion and Future Work}\label{se:discussion}
A new dataset BSCD  consisting of real-world conversation, the service calls, is introduced. 
Human communication is a dynamic process, our eventual goal is to develop a bimodal sentiment analysis engine with the ability to learn the temporal interaction sentiment patterns among conversation parties. 
In the process of fusion, we have approached the study of audio sentiment analysis from an angle that is somewhat different from most people's. 

Future research will concentrate on evaluations using larger data sets, exploring more acoustic feature relevance analysis, and striving to improve the decision-level fusion process. 

A call is constituent of a group of utterances that have contextual dependencies among them. However, in our semi-supervised learning annotation pipeline, about half of the segments in calls are discarded. Therefore the interdependent modeling is out of the scope of this paper and we include it as future work.

        
    
    \bibliography{CallSentimentAnalysis}		
    
    

\end{document}